\def\year{2021}\relax
\documentclass[letterpaper]{article} 
\usepackage{aaai21}  
\usepackage{times}  
\usepackage{helvet} 
\usepackage{courier}  
\usepackage[hyphens]{url}  
\usepackage{graphicx} 
\usepackage{svg} 
\usepackage{natbib}  
\usepackage{caption} 
\usepackage[switch]{lineno}
\usepackage{subcaption}
\usepackage{multirow}
\usepackage{adjustbox}
\usepackage{amssymb}
\usepackage{amsmath}
\usepackage[noend]{algorithmic}
\usepackage[ruled,vlined]{algorithm2e}
\usepackage{xcolor}

\newcommand{\taskset}{\{\tau\}}

\newcommand{\mytitle}{Fixed Priority Global Scheduling from a Deep Learning Perspective\\ Extended Abstract}

\title{\mytitle}
\author {
    Hyunsung Lee\textsuperscript{\rm 1}, Michaël Wang\textsuperscript{\rm 2}, Honguk Woo\textsuperscript{\rm 1}\\
}
\affiliations {
    \textsuperscript{\rm 1} Sungkyunkwan University, South Korea \\ 
    \textsuperscript{\rm 2} Télécom SudParis, France \\ 
    hyunsung.lee@skku.edu,  michael.wang@telecom-sudparis.org,  hwoo@skku.edu
}

\begin{document}
\maketitle

\begin{abstract}
Deep Learning has been recently recognized as one of the feasible solutions to effectively address combinatorial optimization problems, which are often considered important yet challenging in various research domains. In this work, we first present how to adopt Deep Learning for real-time task scheduling through our preliminary work upon fixed priority global scheduling (FPGS) problems.
We then briefly discuss possible generalizations of Deep Learning adoption for several realistic and complicated FPGS scenarios, e.g., scheduling tasks with dependency, mixed-criticality task scheduling. We believe that there are many opportunities for leveraging advanced Deep Learning technologies to improve the quality of scheduling in various system configurations and problem scenarios. 
\end{abstract}

\section{Introduction}
Deep Learning has become a feasible approach to solve complex problems dealing with combinatorial optimization such as Traveling Salesperson problem (TSP) and Routing problems. Upon the recent success of adopting Deep Learning on combinatorial optimization, we showed how to adopt DL-based approaches for tackling an elementary Fixed Priority Global Scheduling (FPGS) problem for large tasksets in~\cite{lee2020panda}. 
%
In FPGS, tasks with time constraints are known \textit{a priori}, and their instances arise \textit{repeatedly} upon a system. Each task is assigned a unique priority offline, and repeated tasks are scheduled online based on task priorities. This pattern is common for embedded systems with fixed priority schedulers~\cite{furst2009autosar}. 
%
In general, FPGS problems are hard to solve for a large taskset due to its combinatorial structure. They have been often dealt with by handcrafted heuristics that require cumbersome efforts in designing and tailoring solutions.

FPGS problems can be cast in sorting tasks in some unknown criterion. 
If there exists a known criterion for comparing tasks, FPGS can fall into a category of  Comparison Sort~\cite{cormen2009introduction} and thus be solved with $O(N\log N)$ where $N$ is the number of tasks in a taskset. 
Most heuristic FPGS algorithms formulate priority assignments as task sorting based on a function of task properties, calculating a score for each task to cast the problem as sorting. For instance, Shortest Job First algorithm sorts tasks according to the execution time of tasks. 
However, except for the case of a single processor platform, such criterion is not known~\cite{liu1973}, and thus heuristics often yield suboptimal priority orders.
Furthermore, without such assumption that tasks can be sorted provided a certain criterion, most FPGS problems are NP-hard, having $N!$ possible candidates, which are infeasible even for a modest-size taskset.

\subsection{Fixed Priority Global Scheduling (FPGS)}
FPGS deals with a predefined set of $N$ tasks where each task $\tau_i$ occurs periodically with its period $T_i$, relative deadline $D_i$, and execution time $C_i$.
Each $\tau_i$ is repeated indefinitely and its invocation may arrive at any time, at least $T_i$ time after the arrival of the previous invocation. We consider a homogeneous single type of resource, a processor. An invocation of $\tau_i$ can only run on a single processor at a time.

For more details of FPGS problem formulation, readers might refer to ~\cite{davis2016review}.

\subsection{Preliminary Work} \label{ssec:prelim}
This section is mostly based on our recent work~\cite{lee2020panda} about a fundamental FPGS problem on a multiple processor platform. 

\subsubsection{Reinforcement learning based approach}

In \cite{vinyals2015pointer}, authors showed that a recurrent neural network, namely Pointer Network, could be used to generate a permutation of inputs, demonstrating its effectiveness for producing permutations that maximize a given objective. 
Later, the work in \cite{bello2016neural} applied the Pointer Network structure for Reinforcement Learning to demonstrate a trainable model without explicit label data but with a score function on the solution that the model makes. They evaluated the approach with TSP. 

\begin{figure}[bht]
    \centering
    \begin{adjustbox}{width=0.8\linewidth}
    \includegraphics[width=\linewidth]{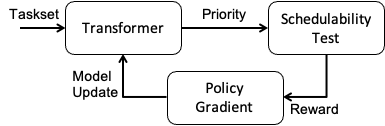}
    \end{adjustbox}
    \caption{High-level overview of Panda: the Transformer-based model generates a priority order (a permutation) for a given taskset. Schedulability Test is used to evaluate rewards for generated permutations. A policy gradient method (e.g., REINFORCE) is used to update model parameters.} \label{fig:panda}
\end{figure}

The work in \cite{kool2018attention} extended this approach with Transformer, showing the wide applicability of Deep Reinforcement Learning to various NP-hard Combinatorial Optimization problems. 
In the same vein of these previous works, our work Panda in \cite{lee2020panda} was the first to apply the idea of solving combinatorial optimization using Deep Learning for FPGS problems. 
The high level concept of Panda using Reinforcement Learning and Transformer for FPGS problems is described in Figure~\ref{fig:panda}.

These works are generally based on either a Recurrent Network or Attention model to build the autoregressive probability distribution over permutations.
\begin{equation}
    Pr\left[\pi|\taskset \right] = \prod Pr\left[\pi_t \vert \pi_1, \pi_2, \dots, \pi_{t-1}, \taskset \right]
\end{equation}
where $\taskset$ is a set of elements to sort, e.g., points in TSP, tasks in FPGS problems. $\pi_t$ is an index of the element $\tau$, which the model chooses at the step $t$. $\pi = [\pi_1, \pi_2, \dots, \pi_N]$ is the output permutation (the order that model generates). 
 
Then, the model can be optimized toward maximizing the policy gradient~\cite{sutton2018reinforcement} equation 
 \begin{equation}
    \mathbb{E} \left[R(\taskset, \pi) Pr\left[\pi|\taskset \right] \right]
 \end{equation} 
where $R(\taskset, \pi)$ denotes the score or reward of output permutations with input instances.
 
\subsubsection{Schedulability test} \label{ssec:schedulability_test}
As described earlier, FPGS problems are NP-hard, requiring examination of $N!$ candidates. This is infeasible for the modest number of tasks $N \simeq 10$. 
To facilitate Reinforcement Learning, we exploited schedulability tests~\cite{Bertogna2009,Bertogna2007,Davis2011,Guan2009} as a reward functions. Schedulability tests provide sufficient conditions by which a given set of tasks is schedulable. Specifically, if a taskset is said to be schedulable by a schedulable test, it is guaranteed to be actually schedulable. However, the inverse condition does not hold.

We used the RTA-LC~\cite{Bertogna2007,Davis2011} as a reward function for a given priority order in Panda. 

\subsubsection{Scalability} \label{ssec:scalability}
\begin{table}[b]
    \centering
    \begin{adjustbox}{width=\linewidth}

    \begin{tabular}{|c||c|l|c|l|c|l|}
    \hline
     $n$-tasks & \multicolumn{2}{c|}{All Perm.(\%)} & \multicolumn{2}{c|}{Random Perm.(\%)} &  
     \multicolumn{2}{c|}{Sched. by DM(\%)} \\ \hline\hline
    4  & \multicolumn{2}{c|}{54.26}                & \multicolumn{2}{c|}{-} & \multicolumn{2}{c|}{94.1} \\ \hline
    6  & \multicolumn{2}{c|}{21.45}                & \multicolumn{2}{c|}{-} & \multicolumn{2}{c|}{92.2} \\ \hline
    8  & \multicolumn{2}{c|}{7.45}                 & \multicolumn{2}{c|}{-}  & \multicolumn{2}{c|}{91.3} \\ \hline
    10 & \multicolumn{2}{c|}{-} & \multicolumn{2}{c|}{2.62}  & \multicolumn{2}{c|}{87.6} \\ \hline
    12 & \multicolumn{2}{c|}{-}                     & \multicolumn{2}{c|}{0.91}  & \multicolumn{2}{c|}{87.3} \\ \hline
    14 & \multicolumn{2}{c|}{-}                     & \multicolumn{2}{c|}{0.27}  & \multicolumn{2}{c|}{87.1} \\ \hline
    16 & \multicolumn{2}{c|}{-}                     & \multicolumn{2}{c|}{0.07}  & \multicolumn{2}{c|}{86.9} \\ \hline
    \end{tabular}
    \end{adjustbox}
    \caption{Ratio of schedulable priority assignments with respect to the size of tasksets ($n$) in \cite{lee2020panda}}
    \label{tbl:feasibleRate}
    \end{table}

As the number of tasks $N$ grows, only the smaller fraction of the permutations from the entire $N!$ candidates are schedulable. In the second and third column of Table~\ref{tbl:feasibleRate}, it is observed that the fraction of schedulable orders decreases exponentially along with larger $N$. This is especially problematic when a model is initialized, e.g., the sampling distribution might be Uniform distribution. The model barely has an opportunity to get nonzero rewards and gradients. 

In Panda, we addressed this issue through two optimization schemes. First, we exploited Deadline Monotonic heuristic for a model to follow using Inverse Propensity Score (IPS)~\cite{degris2012off}  in the early stage of model training. Secondly, we leveraged the response time analysis (RTA) for individual tasks so as to obtain more frequent reward signals in the entire period of model training. It is possible to calculate the response time of partial priority assigned tasks and the upper bound of interference caused by those tasks, as a task can interfere with only higher priority tasks. This calculation can be incorporated into the Reinforcement Learning formulation as rewards, allowing the model to learn quickly with useful reward signals more frequently generated.

\begin{figure}[hbt]
\begin{subfigure}{0.5\textwidth}
  \centering
  \includegraphics[width=.8\linewidth]{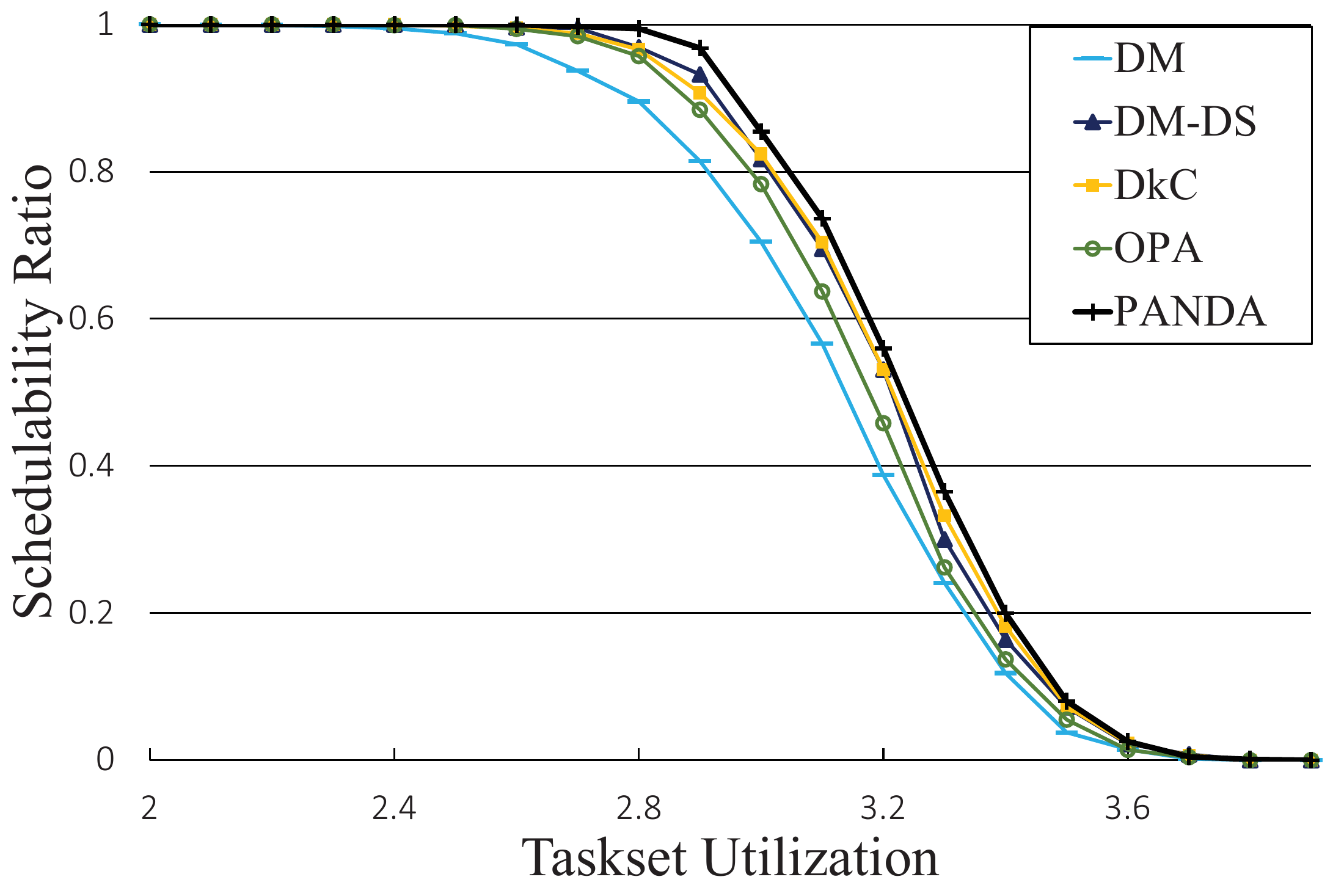}  
  \caption{$m = 4$ and $n = 32$}
  \label{fig:overall-impl-4-32}
\end{subfigure}
\newline
\begin{subfigure}{.5\textwidth}
  \centering
  \includegraphics[width=.8\linewidth]{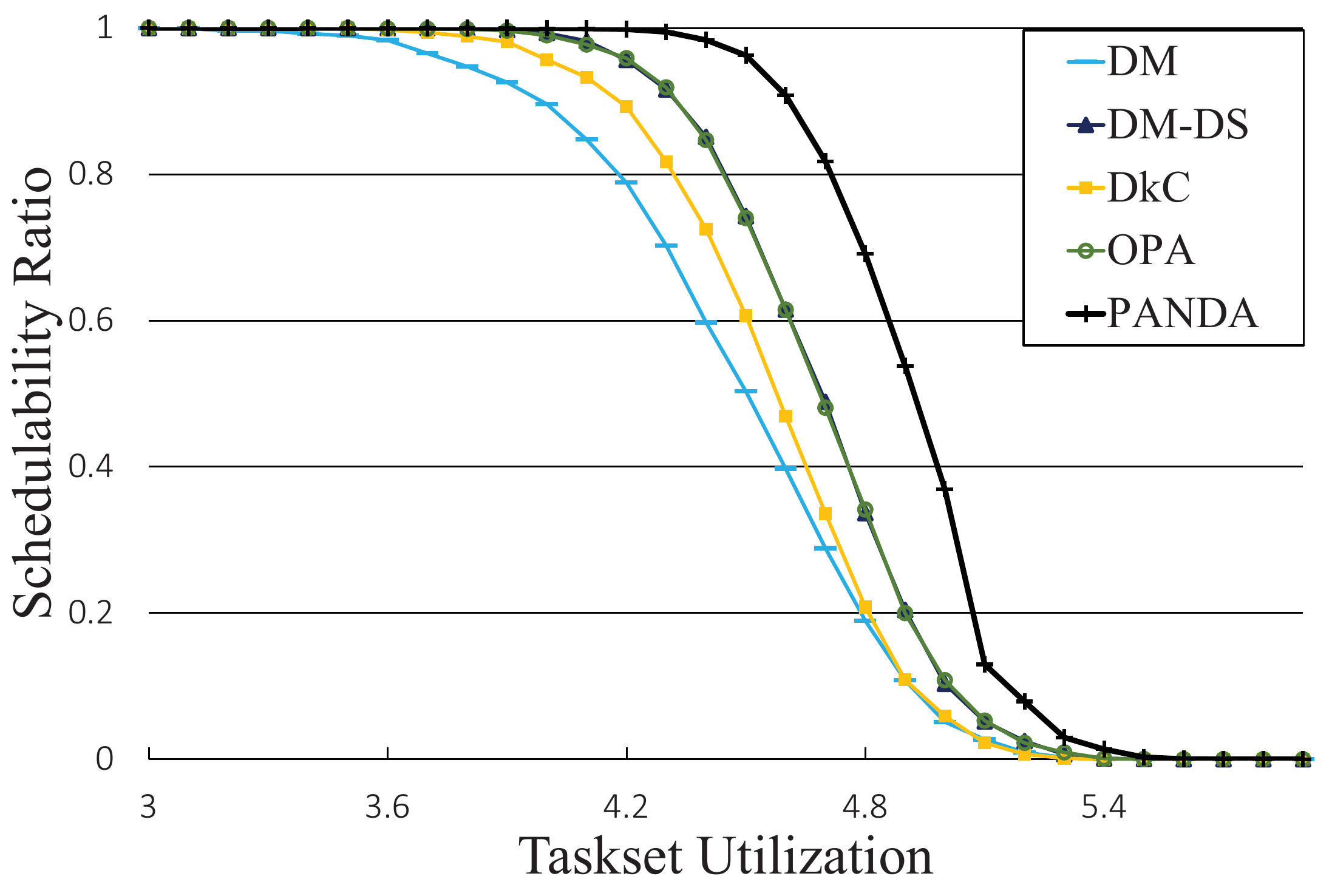}  
  \caption{$m = 6$ and $n = 48$}
  \label{fig:overall-impl=6-48}
\end{subfigure}
\caption{Schedulability ratio with respect to different taskset utilization settings: $m$ denotes the number of processors, and $n$ denotes the number of tasks. For example, in (b), when the system configuration of $48$-sized tasksets $n$ and a $6$-processor platform $m$ is given~\cite{lee2020panda}.}
\label{fig:overall}
\end{figure}

Figure~\ref{fig:overall} shows the benefit of Panda for the preemptive, multiprocessor FPGS problem~\cite{lee2020panda}. The task utilization denotes $\sum_{\tau_i \in \taskset} C_i / T_i$ and the schedulability ratio denotes the portion of schedulable tasks tested using one of the schedulability tests. 
DM, DM-DS, DkC, and OPA are heuristic based baselines. As observed, the performance gap between heuristics and Panda becomes more significant when the number of tasks grows. 


\section{Problem Generalization} \label{sec:generalization}

FPGS has many important problem variants. Here we focus on two specific variants and discuss them, while there can be many different problem settings ~\cite{bril2007worst,vestal2007preemptive}. 
We believe that one of the significant benefits of adopting a Deep Learning model for FPGS problems is that the model can learn competitive heuristic rules without much customization, allowing a unified way to deal with several variants of FPGS.

\subsection{Precedence Condition}

\begin{figure}[ht]
\centering
\includegraphics[width=0.7\linewidth]{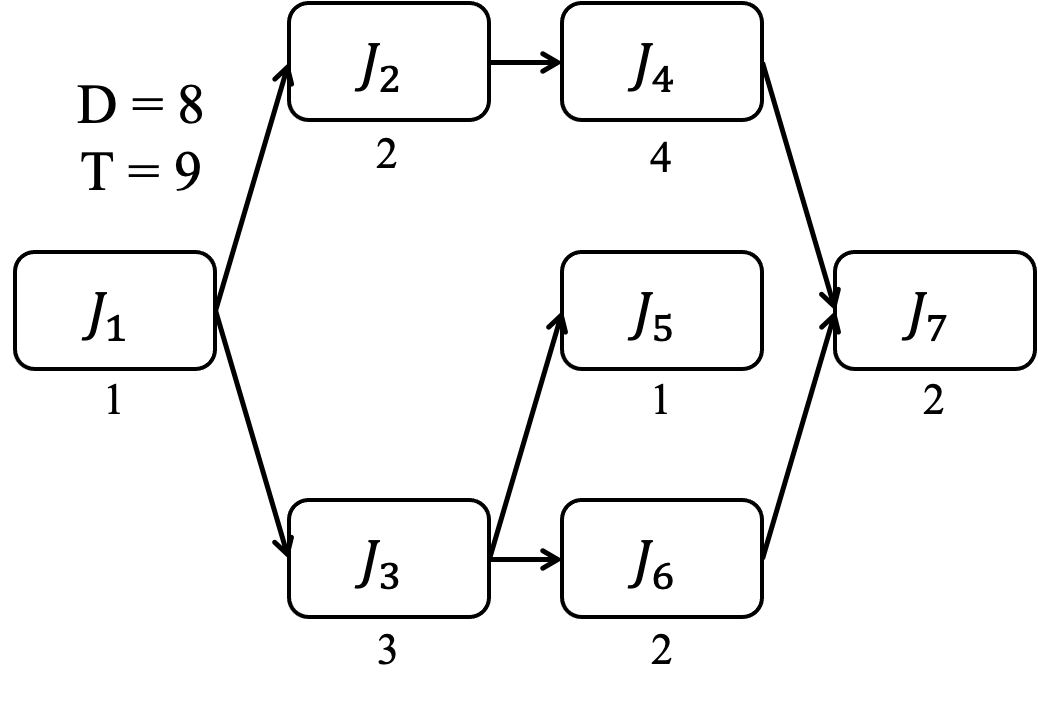}
\caption{An example of a task with precedence conditions: deadline $D$ and period $T$ of the task are on the upper left of the figure. The number below each subtask denotes the execution time of the subtask.} \label{fig:dag}
\end{figure}

One variant for FPGS generalization is that a task can be decomposed into several subtasks with precedence conditions~\cite{baruah2012generalized}. It is related to an efficient resource usage pattern, allowing some fraction of a task to run in parallel on multiple processors and providing  better scheduling opportunities and processor utilization.

We represent a task, its subtasks, and their constraints in a direct acyclic graph (DAG). Each task has subtasks, which are represented by a DAG $G = (V, E)$, as shown in Figure~\ref{fig:dag}. $V$ denotes subtasks for the task and $e_{ij} \in E$ denotes a precedence constraint such that subtask $i$ precedes subtask $j$. An invocation of the task terminates when all of its subtasks are completed. 

Graph Neural Networks can provide an elegant way to handle the complex structure of the FPGS problem with precedence conditions. Specifically, it is possible to construct a Neural Network model with Edge Conditional Convolution layers (ECC)~\cite{simonovsky2017dynamic} to handle tasks represented in DAG. In ECC, the embedding for each node in a graph is learned by iteration over layers. For each node $i$ in the directed graph $G$, embeddings $h_i$ is generated by:
\begin{equation}
    h^{l}_i = \Theta^{l} h^{l}_i + \sum_{j \in \mathcal{N}(i) }\mathbf{k}(a_{ji}) h^{l-1}_i
\end{equation}
where $\mathcal{N}(i)$ denotes nodes which have an edge toward $i$, and $a_{ji}$ denotes the weight of an edge from $j$ to $i$. $\mathbf{k}$ and $\Theta$ at each layer $l$ are learnable parameters. We can represent the graph using linear combinations of nodes, similar to~\cite{li2015gated}:
\begin{equation}
    h_G =  \sum_{i \in G} \frac{w_i}{\sum_j w_j}h_i
\end{equation}
where $w_i$ is given by $\exp(w^Th_i)$. With this representation for nodes (subtasks) and graphs (tasks), we can construct a model to generate priorities for a taskset.

\subsection{Mixed Criticality Systems}
Mixed Criticality specifies several criticalities or levels in a system~\cite{baruah2008schedulability}. At each level, the properties of a task such as execution time and deadline can vary. When a system is in an emergency, it may stop executing less critical tasks to allocate more resources on more critical and emergent tasks. The system might also focus on the Quality of Service regardless of criticality in normal situations. 

Suppose we have a taskset $\taskset$ and $K$ levels for a mixed criticality system. For each level $k$, a task $\tau_i$ may be included or not in the $\taskset_k$. We give each task a unique priority such that scheduling with the taskset runs without violation at any level $k$.

\subsection{Estimating Schedulability Test}
As a problem becomes complex, an algorithm to check if a taskset is deemed schedulable with a priority order becomes computationally intensive. Even there exists a schedulability test for some FPGS variants, for complex systems, the schedulability test can take up to minutes to hours~\cite{nasri2017exact, nasri2019response}. 

Furthermore, it is common that there can be no efficient schedulability tests for some FPGS problems. In that case, it is hard to establish a reliable signal or reward function for priority orders that a model generates. Instead,  approximation methods for schedulability tests can be explored. This is intensively being investigated in the field of Neural Architecture Search (NAS)~\cite{zoph2016neural, elsken2018neural}, which shares a similar problem structure with what we discussed here, i.e., the evaluation of a reward function is non-trivial. Recent research also showed that Deep Learning models can learn to classify structured graph data with a few labeled samples~\cite{chauhan2020few}. 

\section{Conclusion}
In this paper, we revisited Fixed Priority Global Scheduling (FPGS) problems from a perspective of adopting Deep Learning for them. We presented a way to adopt Deep Reinforcement Learning for a straightforward yet fundamental FPGS problem where a set of independent tasks with time constraints is known a priori for multi-processor real-time scheduling. We also briefly discussed two possible FPGS problem variants and the difficulties toward problem generalization. We believe the difficulties can be handled by recent advancements of Deep Learning technologies and hope our work draws attention to the adoption of those technologies for FPGS and other scheduling problems.

\section{Acknowledgement}
This work was supported in part by the ICT Creative Consilience
program under Grant IITP-2020-0-01821 supervised by the IITP (Institute for Information \& communications Technology Planning \&
Evaluation), in part by DNA+ Drone Technology Development Program under Grant 2020M3C1C2A01080819 through the
National Research Foundation of Korea (NRF) funded by the Ministry of Science and ICT (MSIT), and in part by Kakao i Research Supporting Program.

\bibliography{main}
\end{document}